\pdfoutput=1

\documentclass[11pt]{article}

\usepackage{acl}

\usepackage{times}
\usepackage{latexsym}
\usepackage{graphics}
\usepackage{graphicx}
\usepackage{array}
\usepackage{amsmath}
\usepackage{float}
\usepackage{hyperref}
\usepackage{booktabs}
\usepackage{amssymb}

\usepackage[T1]{fontenc}

\usepackage[utf8]{inputenc}

\usepackage{microtype}

\usepackage{inconsolata}

\title{AccentFold: A Journey through African Accents for Zero-Shot ASR Adaptation to Target Accents}

\author{\bf $^\star$Abraham Owodunni$^{1,*}$\quad
  \bf $^\star$Aditya Yadavalli$^{2,*}$ \quad
  \bf  $^\star$Chris Emezue$^{3,4,*}$ \quad
  \bf $^\star$Tobi Olatunji$^{1,*}$ \quad \\
  \bf Clinton Mbataku$^{5,*}$ \\
  $^*$ Masakhane \quad $^1$ Intron Health \quad $^2$ Karya \quad $^3$ Mila Quebec AI Institute \quad $^4$ Lanfrica \\
  $^5$  AI Saturdays Lagos \\
  \tt abraham@intron.io
}

\usepackage{lipsum}

\newcommand\blfootnote[1]{%
  \begingroup
  \renewcommand\thefootnote{}\footnote{#1}%
  \addtocounter{footnote}{-1}%
  \endgroup
}

\begin{document}
\maketitle
\begin{abstract}
Despite advancements in speech recognition, accented speech remains challenging. While previous approaches have focused on modeling techniques or creating accented speech datasets, gathering sufficient data for the multitude of accents, particularly in the African context, remains impractical due to their sheer diversity and associated budget constraints. To address these challenges, we propose \textit{AccentFold}, a method that exploits spatial relationships between learned accent embeddings to improve downstream Automatic Speech Recognition (ASR).  Our exploratory analysis of speech embeddings representing 100+ African accents reveals interesting spatial accent relationships highlighting geographic and genealogical similarities, capturing consistent phonological, and morphological regularities, all learned empirically from speech. Furthermore, we discover accent relationships previously uncharacterized by the Ethnologue. Through empirical evaluation, we demonstrate the effectiveness of AccentFold by showing that, for out-of-distribution (OOD) accents, sampling accent subsets for training based on AccentFold information outperforms strong baselines with a relative WER improvement of 4.6\%. AccentFold presents a promising approach for improving ASR performance on accented speech, particularly in the context of African accents, where data scarcity and budget constraints pose significant challenges. Our findings emphasize the potential of leveraging linguistic relationships to improve zero-shot ASR adaptation to target accents. Please find our code for this work here.\footnote{\url{https://github.com/intron-innovation/accent_folds}}
\blfootnote{$\star$ Authors contributed equally}

\end{abstract}

\section{Introduction}

English language is spoken in 88 countries and territories as either an official, administrative, or cultural language, estimated at over 2 billion speakers with non-native speakers outnumbering native speakers by a ratio of 3:1. 





Despite considerable advancements, automatic speech recognition (ASR) technology still faces challenges with accented speech \cite{yadavalli-etal-2022-exploring, szalay22-interspeech, sanabria23edacc}.
Speakers whose first language (L1) is not English have high word error rate for their audio samples \cite{dichristofano2022performance}. \citet{koenecke2020racial} showed that existing ASR systems struggle with speakers of African American Vernacular English (AAVE) when compared with speech from rural White Californians. 

The dominant methods for improving speech recognition for accented speech have conventionally involved modeling techniques and algorithmic enhancements such as multitask learning \cite{Jain2018, Zhang2021E2EBasedML, yadavalli22_interspeech, 8461886}, domain adversarial training \cite{Feng2021QuantifyingBI, Li2021AccentRobustAS}, active learning \cite{Chellapriyadharshini_2018}, and weak supervision \cite{Khandelwal2020BlackboxAO}. Despite some progress in ASR performance, performance still degrades significantly for out-of-distribution (OOD) accents, making the application of these techniques in real-world scenarios challenging. To enhance generalizability, datasets that incorporate accented speech have been developed  \cite{commonvoice, sanabria23edacc}. However, given the sheer number of accents, it is currently infeasible to obtain a sufficient amount of data that comprehensively covers each distinct accent. 

In contrast, there has been a relatively smaller focus on exploring linguistic aspects, accent relationships, and harnessing that knowledge to enhance ASR performance. Previous research in language modeling \cite{nzeyimana2022kinyabert}, intent classification \cite{sharma2021leveraging} and speech recognition \cite{8461972,li2021accentrobust, Jain2023HowDP} have demonstrated that incorporating linguistic information in NLP tasks generally yields downstream improvements, especially for languages with limited resources and restricted data availability -- a situation pertinent to African languages. Consequently, we opine that a deeper understanding of geographical and linguistic similarities, encompassing syntactic, phonological, and morphological aspects, among different accents can potentially enhance ASR for accented speech.

We believe embeddings offer a principled and quantitative approach to investigate linguistic, geographic and other global connections \cite{mikolov2013distributed,garg2018word}, and form the framework of our paper. Our contribution involves the development of AccentFold, a network of learned accent embeddings through which we explore possible linguistic and geographic relationships among African accents. We report the insights from our linguistic analysis in Section \ref{sec:ling-analyis}.

By conducting empirical analysis, we demonstrate the informative nature and practical significance of the the accent folds. Concretely, in Section \ref{sec:exp}, we show that for a given target OOD accent, fine-tuning on a dataset generated from a subset of accents obtained through AccentFold leads to improved performance compared to strong baselines. 

\section{Related Work}

Using existing state-of-art pre-trained models to probe for linguistic information and using that to improve models' performance has gained interest in the community recently. \citet{prasad-jyothi-2020-accents} use various probing techniques on the DeepSpeech 2 model \cite{Amodei2015DeepS2}. They find that first few layers encode most of the accent related information. \citet{bartelds-wieling-2022-quantifying} quantify language variation in Dutch using a combination of XLS-53 \cite{Conneau2020UnsupervisedCR} embeddings and Dynamic Time Warping \cite{1163055}. They show that this leads to a Dutch dialect identification system that is better than a system dependent on the phonetic transcriptions with just six seconds of speech. Thus, proving that pre-trained models such as the one proposed by \citet{Conneau2020UnsupervisedCR} indeed capture rich linguistic information in their representations. \citet{Jain2018, Li2021AccentRobustAS} extract accent embeddings learnt from a separate network and input those embeddings along with other features. They show that this leads to a superior accented ASR model. Our work is most closely related to \cite{kothawade-etal-2023-ditto}, where the authors explore various statistical methods such as \textit{Submodular Mutual Information} in combination with hand-crafted features to select a subset of data to improve accented ASR. Our work differs from previous works in two important ways (1) we take a different approach and use the extracted accent embeddings from a pre-trained model to decide what subset of data to use to build an ASR that performs the best on a target accent in a cost-effective manner (2) we do this at a much larger scale of 41 African English accents. Note that the previous highest was 21 English accents by \citet{Li2021AccentRobustAS}.

\section{AccentFold}
This section outlines the procedures involved in the development of AccentFold.
\subsection{The Dataset}

\begin{figure}[h!]
    \centering
    \includegraphics[width=8cm]{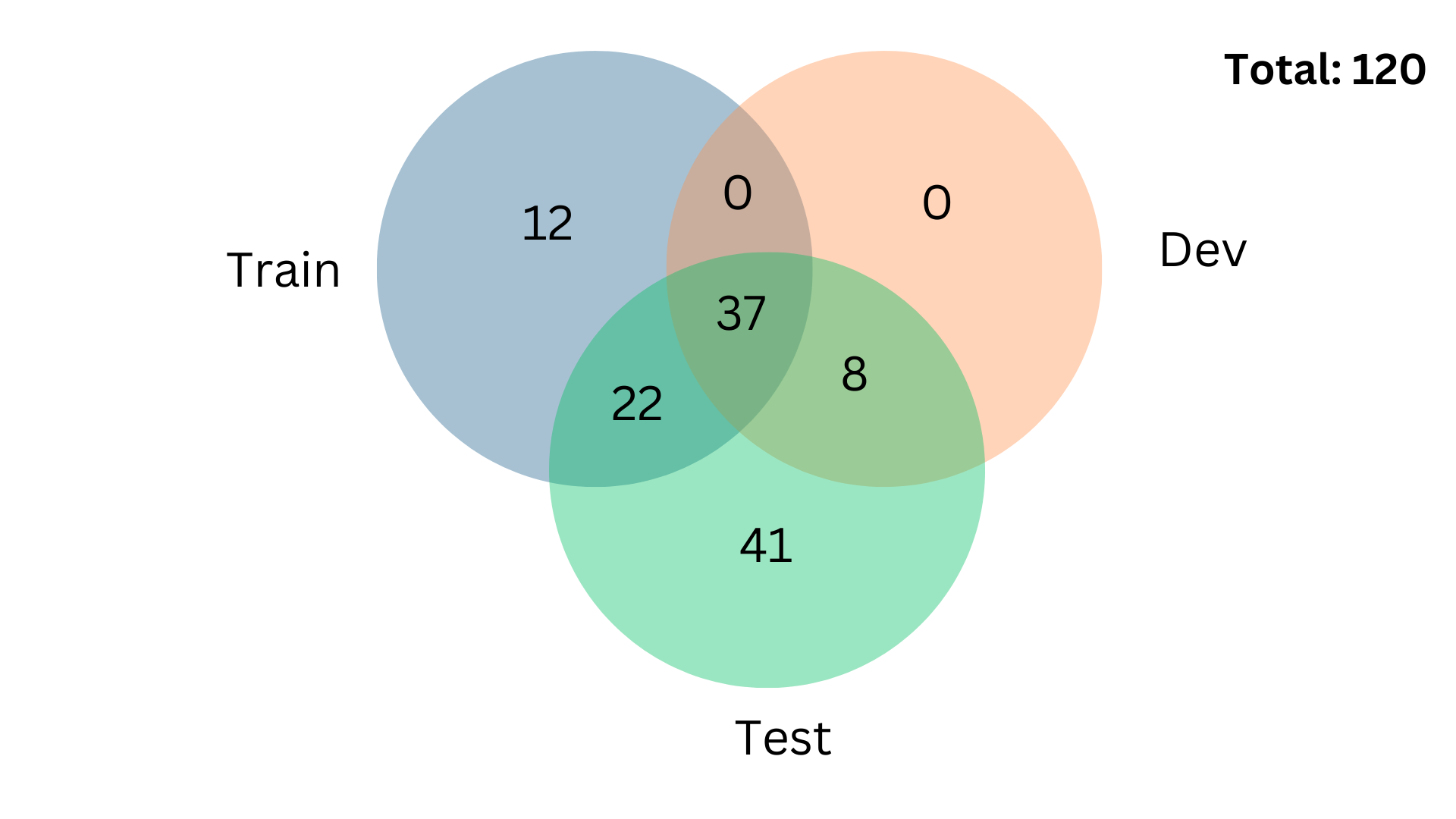}
    \caption{Venn diagram of the accent splits}
    \label{fig:venn-diagram}
\end{figure}

We use the Afrispeech-200 dataset \cite{olatunji2023afrispeech200} for this work, an accented Pan-African speech corpus with over 200 hours of audio recording, 120 accents, 2463 unique speakers, 57\% female, from 13 countries for clinical and general domain ASR. To the best of our knowledge, it is the most diverse collection of African accents and is thus the focus of our work. Table \ref{tab:dataset_stats} shows the statistics of the full dataset and \autoref{syn-stats} focuses on the accentual statistics of the Afrispeech-200 dataset. With 120 accents, the dataset covers a wide range of African accents. The entire dataset can be split, in terms of accents, into 71 accents in the train set, 45 accents in the dev set and 108 accents in the test set, of which 41 accents are only present in the test set (see \autoref{fig:venn-diagram}). The presence of unique accents in the test split enables us to model them as Out Of Distribution (OOD) accents: a situation beneficial for evaluating how well our work generalizes to unseen accents.

\begin{table}
\small
\centering
\begin{tabular}{cc}
\toprule
\textbf{Speaker Gender Ratios} & \textbf{No. of Utterances \%} \\
\midrule
Female & 57.11\%  \\
Male & 42.41\%  \\
Other/Unknown & 0.48\% \\
\midrule
\textbf{Speaker Age Groups} & \textbf{No. of Utterances \%} \\
\midrule
<18yrs & 1,264 (1.88\%) \\
19-25 & 36,728 (54.58\%)  \\
26-40 & 18,366 (27.29\%)  \\
41-55 & 10,374 (15.42\%) \\
>56yrs & 563 (0.84\%)  \\
\midrule
\textbf{Domain} & \textbf{No. of Utterances \%} \\
\midrule
Clinical & 41,765 (61.80\%)  \\
General & 25,812 (38.20\%) \\
\bottomrule
\end{tabular}
\caption{Afrispeech-200 Dataset statistics}
\label{tab:dataset_stats}
\end{table}

\subsection{Creating AccentFold}
\paragraph{Obtaining and visualizing accent embeddings:}


AccentFold is made up of learned accent embeddings. To create the embeddings, we follow the work of \citet{mtl}. This is a multitask learning model (MTL) on top of a pre-trained XLS-R model \cite{Conneau2020UnsupervisedCR}. 
The MTL model contains a shared encoder with three heads : (1) ASR head (2) Accent classification head, and (3) Domain classification head. The \textbf{accent classification} head predicts over 71 accents while the \textbf{Domain classification} head predicts (binary) if a sample is from the clinical or general domain. The ASR head is trained with the Connectionist Temporal Classification (CTC) loss \cite{10.1145/1143844.1143891} using the same hyperparameters as \citet{Conneau2020UnsupervisedCR}. For the domain and accent heads, we perform mean pooling on the encoder output and pass this to the dense layers in each corresponding head. The \textbf{accent classification} head predicts over 71 accents with cross-entropy loss. Extreme class imbalance further makes the task challenging. Therefore, we add a dense layer to our accent classification head to model this complexity.  \textbf{Domain classification} uses a single dense layer with binary cross-entropy loss. 
The 3 tasks are jointly optimized as follows:

 $$ L_{MTL} = 0.7 p_{ctc}(y|x) + 0.2 p_{acc}(a|x) + 0.1 p_{dom}(d|x)$$


We found the above relative weights to give us the best results. For all the experiments, we train the models with a batch size of 16 for 10 epochs. Following \citet{Conneau2020UnsupervisedCR}, we use Adam optimizer \cite{kingma2014adam} where the learning rate is warmed up for the first 10\% of updates to a peak of 3e-4, and then linearly
decayed over a total of 30,740 updates. We use Hugginface Transformers to implement this \cite{wolf2020huggingfaces}.  

We train this model on the AfriSpeech-200 corpus \cite{olatunji2023afrispeech200}. We then extract internal representations of the last Transformer layer in the shared encoder model and use these as our \textit{AccentFold} embeddings. 
For all samples for a given accent, we run inference using the MTL model and obtain corresponding \textit{AccentFold} embeddings. For a given set of accent embeddings, we create a centroid represented by its element-wise medians. We select the median over the mean because of its robustness to outliers. 

To visualize these embeddings we use t-distributed stochastic neighbor embedding (t-SNE) \cite{Maaten2008VisualizingDU} with a perplexity of 30 and early aggregation of 12 to transform the embeddings to 2 dimensions. 
Initially, we apply the t-SNE transformation to the entire Afrispeech dataset and create plots based on the resulting two-dimensional embeddings. This step enables us to visualize the overall structure and patterns present in the dataset. Subsequently, we repeat the transformation and plotting process specifically for the test split of the dataset. This evaluation allows us to determine if the quality of the t-SNE fitting and transformation extends to samples with unseen accents. 

\begin{figure*}[h!]
\includegraphics[width=\textwidth]{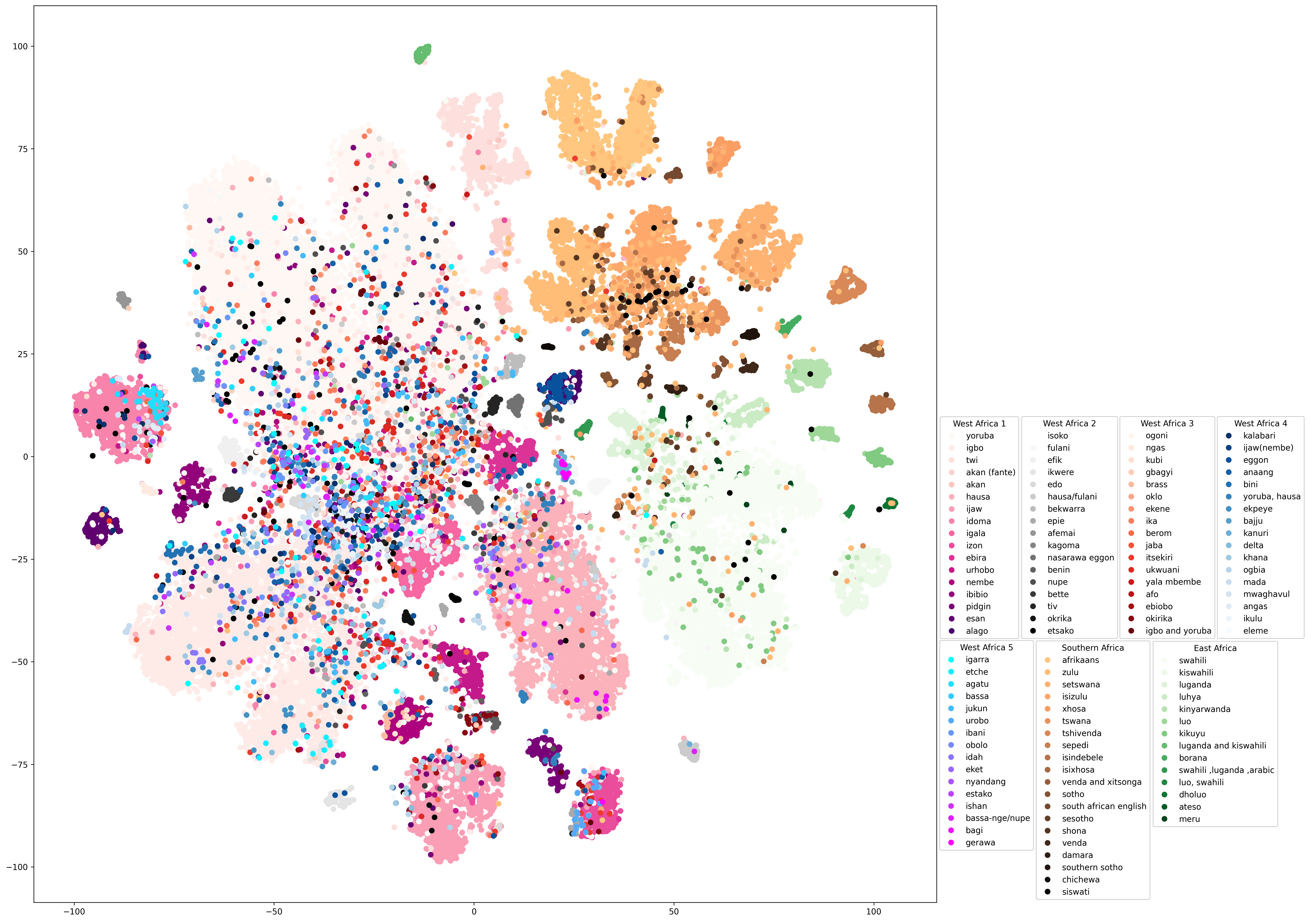} %
\caption{t-SNE visualization of the learned accent embeddings in AccentFold: embeddings of the entire Afrispeech-200 data. In this figure, each accent is encoded with one color. We use the color transparency to differentiate the accents, while the color categories represent the geographical region.}
\label{afrispeech_train_dev_test_by_acccent}
\end{figure*}

\section{What information does AccentFold capture?}
\label{sec:ling-analyis}
\begin{figure}[h]
\includegraphics[width=8cm]{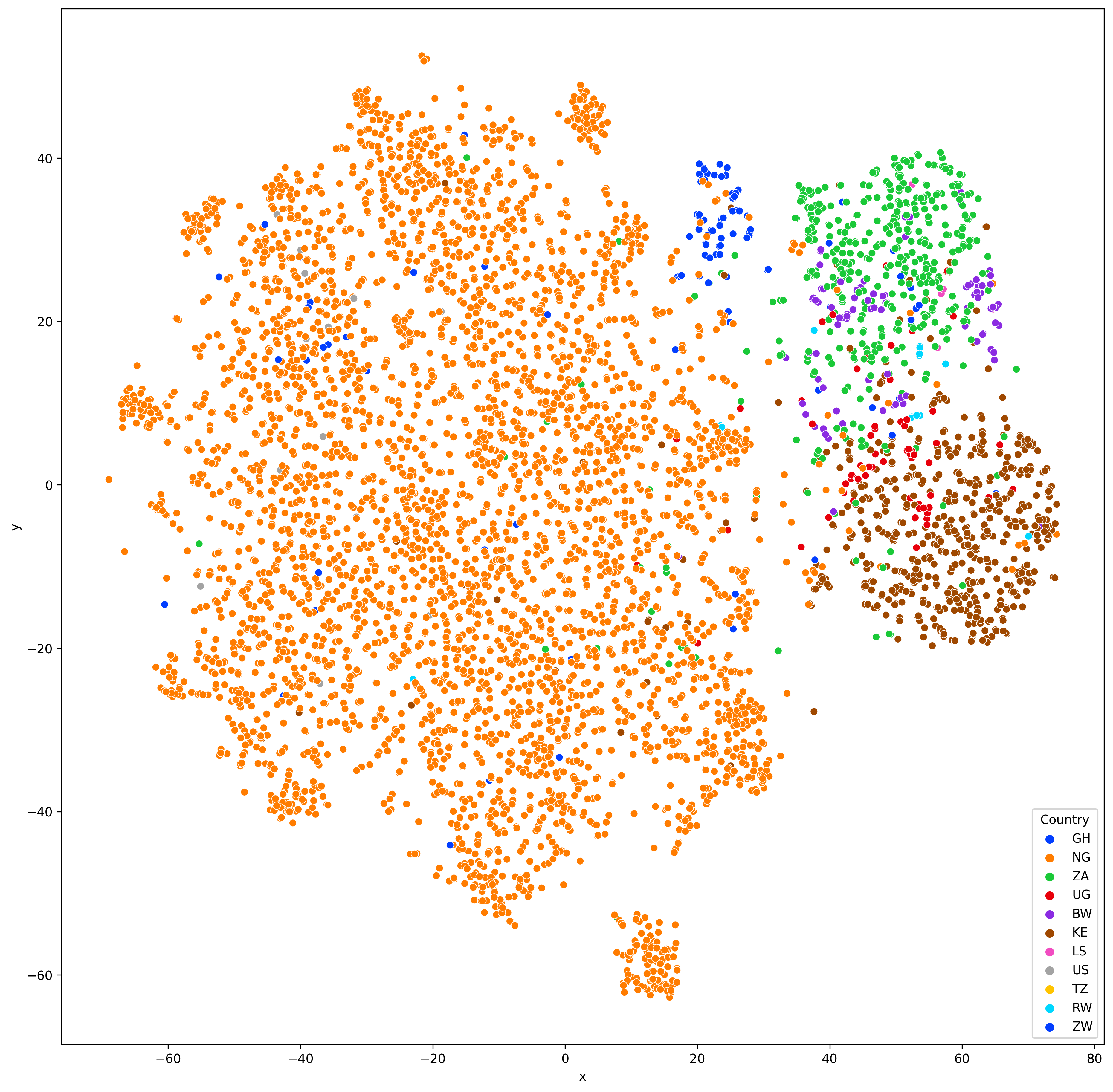} %
\caption{t-SNE visualization of embeddings by country from the Afrispeech test split.} 
\label{afrispeech_by_country}
\end{figure}

\begin{figure}[t]
\includegraphics[width=8cm]{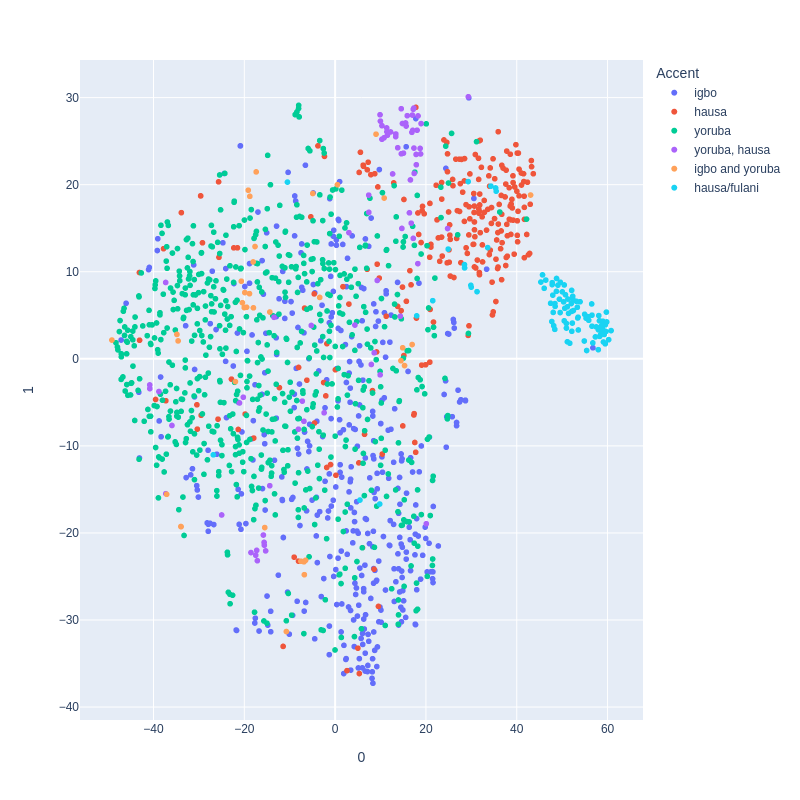} %
\caption{Analysis of Dual Accents}
\label{afrispeech_dual_accents}
\end{figure}

In this section, we delve into an exploratory analysis of the t-SNE visualizations for all the accents in AccentFold. Our aim is to gain a deep understanding of the intricate connections and patterns that emerge among these diverse accents. The t-SNE visualizations of the accent in AccentFold can be found in Figures 
\ref{afrispeech_train_dev_test_by_acccent}, 
\ref{afrispeech_by_country}, 
\ref{afrispeech_dual_accents}. We also present some more Figures (\ref{afrispeech_test_all_data_by_accent}, \ref{afrispeech_test_all_data_by_families}, \ref{afrispeech_train_dev_test_by_family}, \ref{afrispeech_by_region}) in the Appendix.

\paragraph{Language Families:} 

Figure \ref{afrispeech_train_dev_test_by_family} presents a t-SNE visualization of the learned accent embeddings, where color coding is utilized to distinguish language families, and varying levels of transparency ensure distinct colors for each accent. Each point in the figure corresponds to an accent embedding obtained through AccentFold, allowing us to convey two pieces of information: the distribution of accents and their respective language families.


Through an exploratory analysis of Figure \ref{afrispeech_train_dev_test_by_family}, we observe that the accent embeddings tend to group together (forming what we refer to as ``accent folds'') based on language family similarities. Language families represent the genetic connections between languages, as they consist of languages that descended from a common ancestor \cite{linguisticsnetworkIntroLanguage}. These language families exhibit syntactic, phonological, and morphological relationships \cite{doi:10.1146/annurev-linguistics-011718-011842}. Based on these observations, we hypothesize that AccentFold captures linguistic regularities within accents.

\paragraph{Geographically Consistent Clusters:}
Although the majority of the data comes from Nigeria, \autoref{afrispeech_by_country} plots all test samples with their country labels showing spatial relationships between countries. The t-SNE plots generally align with geographical disposition, accents from Nigeria (Orange) are closer in vector space to Ghana (blue) but further from Kenya, Uganda, Rwanda, and South Africa likely reflecting the distinct languages spoken across these countries. However, where similar languages (e.g. Swahili) are spoken across countries (e.g. Botswana and South Africa), the spatial distinction is less apparent. Uganda, Kenya, and Tanzania cluster together while Botswana and South Africa cluster together and Rwandan embeddings fall between both regions. This demonstrates that the learned embeddings do encode some geographical information extracted entirely from speech and accent labels.

\paragraph{Accent disposition:}
In Figure \ref{afrispeech_test_all_data_by_accent}, Ghanaian accents - Twi and Akan (Fante), cluster closer together and are distinct from Nigerian neighbors. South African accents Zulu, Afrikaans, and Tswana cluster together. Similarly, Kinyarwanda, Luganda, Luganda, Swahili, Luhya and other East African accents cluster together. In Nigeria, Northern accents Hausa and Fulani cluster together and are closer to middle belt accents than South-Eastern and South-Western Nigerian accents. Accents spoken in South-Eastern Nigeria, which make up the majority of West African accents in this dataset, represent the collection of embeddings with indistinguishable margins, representing the close relationship between these accents.

\paragraph{Peripheral West African Clusters:}
Figure \ref{afrispeech_by_country} shows a distinct pattern in the Nigerian accents. There are 10 distinct peripheral subclusters surrounding a more homogenous core. These may represent accents with very distinct linguistic or tonal characteristics from various parts of the country. Some of these accents include  Okirika, Bajju, Brass, Agatu, Eggon, Mada, Ikulu Hausa and Urobo.


\paragraph{Dual Accents:}
Figure \ref{afrispeech_dual_accents} shows a really interesting phenomenon with speakers with self-reported dual accents. Sample embeddings for dual accents "Igbo and Yoruba" (orange) fall between the Igbo (blue) and Yoruba (green) clusters. Although Yoruba (green) and Hausa (red) are very distinct accents, speakers with dual accents (purple) fall somewhat between both clusters. This trend is consistent with Yoruba/Hausa and Hausa/Fulani accents.

\subsection{Contrasting with the Ethnologue}
\label{sec:ethnologue-challenge}
According to Ethnologue \cite{campbell2008ethnologue} there are 7,151 living human languages distributed in 142 different language families, 6 of which are assigned to Africa, based on historically accepted language ancestry. Although the empirically learned  embeddings generally support this classification, they reveal 2 interesting possibilities that remain uncharacterized by the Ethnologue.

\paragraph{Kwa-Bantu Relationship:}
Although the Ghanaian Kwa languages are traditionally separated from the Bantu languages in South Africa and are geographically very distant, our embeddings suggest they may be more similar than earlier proposed and possibly share similar ancestry. This line of reasoning is supported by \citet{guldemann2018historical} reclassification of African languages.


\paragraph{Niger-Congo Subfamilies.}
Although there have been attempts to better categorize the large Niger-Congo family, \citet{guldemann2018historical}'s work, based on basic classificatory units and genealogical relations, rethinks traditional classification. The spatial disposition shown in Figure \ref{afrispeech_test_all_data_by_families} also suggests possible sub-families based on speech representations empirically learned by optimizing the MTL objective function.

\subsection{Accent Normalization and Re-identification}
User reported accents are sometimes noisy. In the Afrispeech dataset, we encountered 4 strange accent labels where their groupings shed more light on possible true accent labels. 11 speakers located in Nigeria reported their accent as ``English''. Although the centroid for this group is closest to the ``Berom" accent, all samples for this group fall within clusters occupied by speakers from Southeastern Nigeria. Another group of 20 speakers reported a ``pidgin'' accent. Embedding for speech for speakers are nearest to clusters from Ijaw, Delta, Edo, and other Nigerian accents where pidgin accent is prevalent. 2 speakers self-identified their accents as ``South African English''. However embeddings are closest to Afrikaans speakers. Embeddings for a group of ``Portugese'' speakers located in South Africa also fall very close to Zulu and Tswana, both south African accents. Embedding/Accent distances were also very valuable with normalizing dialects or misspelled accents for example ``luo'' and ``dholuo'', ``Twi'' and ``Akan'', ``kiswahili'' and ``swahili'' and many others.



\section{Empirical study of AccentFold}
\label{sec:exp}
\subsection{Problem Formulation}
In this empirical study, we set out to understand how informative the accent folds are for accent-level zero shot ASR performance. To achieve this, we designed our experimental task as follows: 
Assume we have the below oracle data set generator:
\begin{equation}
\label{eqn1}
    \mathbb{F(a_{k})} \longrightarrow \{(x_{i},y_{i})\}_{i=1}^{N_{k}},
\end{equation}
such that when $\mathbb{F}$ is given an accent $a_{k} \in A:=\{a_{1},a_{2},a_{3},...,a_{n}\}$, it returns a data set of $N_{k}$ audio-text pairs where the audio samples are from speakers of accent $a_{k}$. $A$ is a finite set of possible accents from which the generator can give us data samples. Also, $N_{k}$ varies for each accent $a_{k}$. We have a target OOD accent $a_{OOD} \notin A$ for which we want to improve ASR performance. For every given OOD target accent $a_{OOD}$, we can only select $s << n$ accents from $A$, i.e $A_{s} = \{a_{1},...,a_{s}\}$, with which we can obtain data samples from $\mathbb{F}$ and finetune our model. The problem then becomes how to choose $A_{s}$ for a given $a_{OOD}$. 

As a practical example of the problem above, consider a company that wants to improve their speech recognition performance on $a_{OOD}$. They therefore hire recorders with various accents ($A$) to record given texts, but do not have access to recorders with accent $a_{OOD}$ perhaps due to geographical reasons (a company based in the USA would find it difficult to find speakers with \textit{afante} accent). Due to constraints (perhaps budget, time) they can not engage all the recorders in the recording task. So it is imperative to choose which accents to use to create the training dataset for their ASR system. This is an important problem in the real world, where accents are abound and resource constraints are highly limited \cite{aksenova2022accented,hinsvark2021accented}.

The approach we adopt as our baseline is to select $A_{s}$ randomly. AccentFold offers another approach to selecting $A_{s}$: by selecting accents from $A$ that share geographic and linguistic similarities with $a_{OOD}$.

\subsection{Experimental Setup}
For our experimental setup, we interpret the Afrispeech-200 dataset as our oracle dataset and design a function, $\mathbb{F(a_{k})}$, that returns the speech-text samples from Afrispeech-200 which are spoken with accent $a_{k}$. $A$ then represents the distinct set of accents in Afrispeech-200. We visualize in \autoref{fig:venn-diagram} a Venn diagram showing how the accents intersect within the train, test and dev splits.   


\paragraph{Target accents ($a_{OOD}$):}

Based on \autoref{fig:venn-diagram}, we note the presence of 41 accents within the test split that are not found in either the train or dev splits. As a result, we choose these 41 accents to represent our target the out-of-distribution (OOD) accents for our experimental setup. We choose our $s$ to be 20.

\paragraph{Selecting $A_{s}$ and obtaining fine-tuning dataset:}
Our experimental setting is hinged on how we select the accent subset, $A_{s}$, from which the data generator retrieves the fine-tuning dataset will be used. For our first baseline, we implement a random selection of $s$ accents from $A$. Sampling is done uniformly and without replacement. 

For our second baseline (GeoProx), we leverage the real-world geographical proximity of the accents. Concretely speaking, for a given target OOD accent, $a_{OOD}$, we extract its country information and compare this information with that of the other accents in $A$, taking the $s$ accents that are geographically closest to $a_{OOD}$. We leverage the geocoding Python package called \textit{geopy}\footnote{\url{https://github.com/geopy/geopy}} for this process.

With the utilization of AccentFold, we extract the centroids of the accents in $A$, as well as a given OOD accent $a_{OOD}$. Leveraging the vectorial representation of accents, determining their similarities becomes straightforward using the cosine distance metric. Consequently, we compute the cosine similarity between the embedding vector of the OOD target accent and that of each accent in $A$. We subsequently arrange the accents in $A$ in ascending order based on their cosine similarity and select the top $s$ accents, resulting in the formation of $A_{s}$ for a given $a_{OOD}$. We perform this operation for each of the 41 accents in our target accent set.

Then for each $a_{OOD}$ we utilize our data generator to obtain a training dataset $\mathbb{D} =  \{(x_{i},y_{i})\}_{i=1}^{N_{k}} $ of speech-text samples based on accents in $A_{s}$. This dataset is then used for our fine-tuning experiment which is explained in more detail below. 

\begin{figure*}[t!]
    \includegraphics[width=\textwidth]{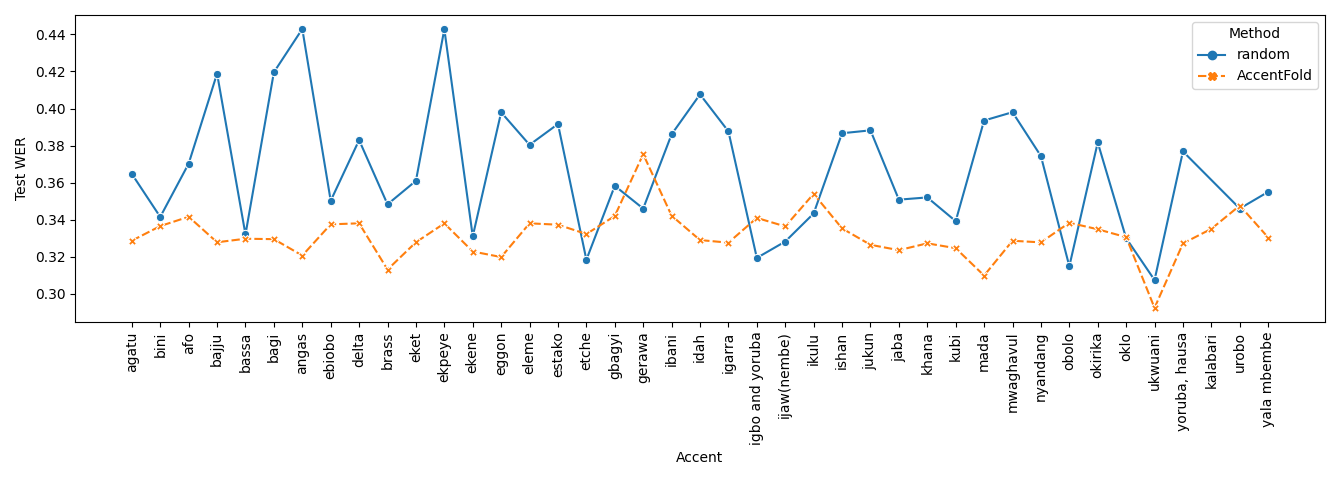}
    \caption{Test WER across all 41 OOD accents. We compare AccentFold with random sampling.}
    \label{fig:test-wer-41}
\end{figure*}

\paragraph{Fine-tuning Details:}
We use a pre-trained XLS-R model \cite{Conneau2020UnsupervisedCR} for our experiments. The XLSR model extends the wav2vec 2.0 \cite{wav2vec2} model to the cross-lingual setting and was trained to acquire cross-lingual speech representations through the utilization of a singular model that is pre-trained using raw speech waveforms from various languages. 
The fact that this model is cross-lingual makes it a good fit for our experiments.

During the fine-tuning of our pre-trained model, we follow the hyperparameter settings of \citet{olatunji2023afrinames}. These include setting the dropout rates for attention and hidden layers to 0.1, while keeping the feature projection dropout at 0.0. We also employ a mask probability of 0.05 and a layerdrop rate of 0.1. Additionally, we enable gradient checkpointing to reduce memory usage. The learning rate is set to 3e-4, with a warm-up period of 1541 steps. The batch sizes for training and validation are 16 and 8, respectively, and we train the model for ten epochs. 

For each of the 41 target accents, we finetune our pre-trained model on its corresponding dataset and evaluate the word error rate on the test set comprising audio samples containing only the target accent. We run all our experiments using a 40GB NVIDIA A100 SXM GPU, which enables parallel use of its GPU nodes.

\textbf{Evaluation procedure: } It is important to note that although the training dataset size $N_{k}$ depends on the target accent $a_{OOD}$ in consideration, the test set used to evaluate all our experiments is fixed: it comprises the samples from the test split of the Afrispeech-200. Using Figure \ref{fig:venn-diagram} the test set are samples from all the 108 accents of the test split. By keeping the test set constant, we can assess the model's performance on our intended accent $a_{OOD}$ in an out-of-distribution (OOD) scenario. This is because the training and development splits do not include any audio-speech samples from these accents. Additionally, this procedure enables us to evaluate the model's capacity to generalize to other accent samples, resulting in a highly resilient evaluation.

\subsection{Results and Discussion}

\begin{table}[htb!]
\caption{Test WER on target OOD accent compared by subset selection using AccentFold, GeoProx, and random sampling. Average and standard deviation are taken over the 41 accents of our target. We also report p-value from a 1-sample, two-sided t-test.}
\label{wer}
\begin{center}
\begin{tabular}{cc}
\toprule
\textbf{Model} & \textbf{Test WER $\downarrow$} \\
\midrule
AccentFold & $\textbf{0.332} \pm \textbf{0.013}$ \\ 
GeoProx & $0.348 \pm 0.007$ \\ 
Random & $0.367 \pm 0.034$ \\ 
\bottomrule
\end{tabular}
\end{center}
\end{table}

\begin{figure}[ht!]
    \centering
    \includegraphics[width=0.5\textwidth]{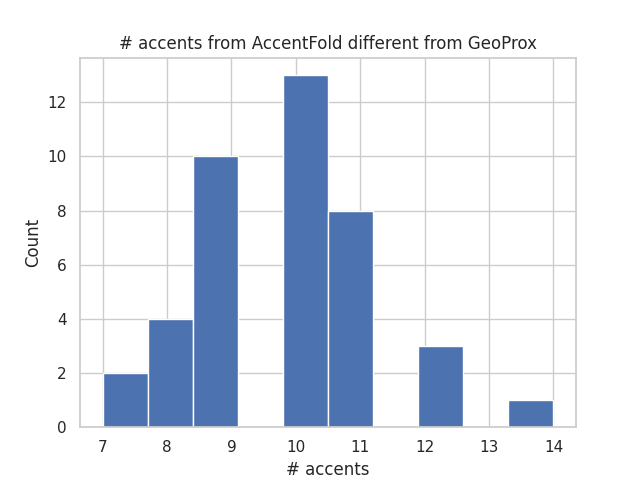}
    \caption{Histogram of number of accents from AccentFold that are non-overlapping with GeoProx.}
    \label{fig:hist-overlap}
\end{figure}

Table \ref{wer} presents the results of a test Word Error Rate (WER) comparison between three different approaches for subset selection: AccentFold, GeoProx, and random sampling. The table displays the average and standard deviation of the WER values over the 41 target OOD accents. The results show that the AccentFold approach achieves the lowest test WER of 0.332 with a standard deviation of 0.013. In contrast, the random sampling approach yields the highest test WER of 0.367 with a larger standard deviation of 0.034. GeoProx, which uses real-world geographical proximity of the accents, performs better than random sampling but still under-performs when compared to AccentFold. To better understand this, we investigate the accents selected by AccentFold and GeoProx and analyse their non-overlapping accents in Figure \ref{fig:hist-overlap}. The histogram reveals that many of the accents selected by AccentFold for any given target OOD accent, $a_{OOD}$, are not necessarily those geographically closest to $a_{OOD}$. This insight suggests that the learned embeddings in AccentFold encompass much more than geographical proximity of accents.   



Figure \ref{fig:test-wer-41} visualizes the test WER obtained by AccentFold and random sampling for each of the 41 accents. We see that in majority of the accents, AccentFold leads to improved performancte in terms of WER compared to random sampling. These findings indicate that AccentFold effectively captures linguistic relationships among accents, allowing for more accurate recognition of the target OOD accent when used to build the fine-tuning dataset. This demonstrates the usefulness of leveraging linguistic information and accent embeddings provided by AccentFold in the context of automatic speech recognition tasks.



\begin{figure}[ht!]
    \centering
    \includegraphics[width=0.5\textwidth]{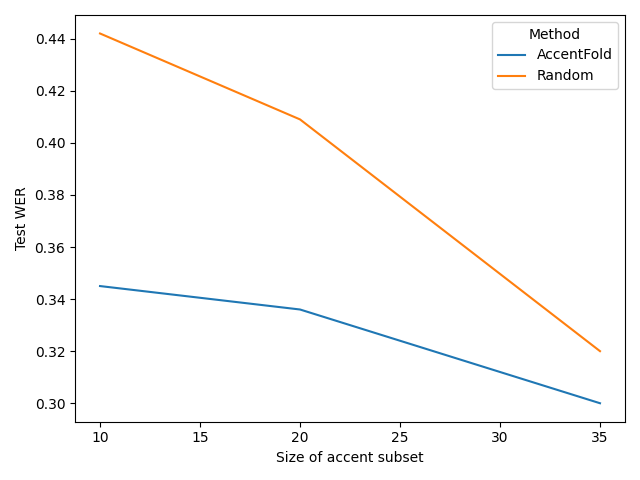}
    \caption{Test WER on Bini accent for different accent subset sizes (different values of $s$ for $A_{s}$).}
    \label{fig:trend-accent-subset}
\end{figure}

We notice a pattern, as shown in Figure \ref{fig:trend-accent-subset}, where increasing the value of $s$, which corresponds to a larger training dataset size $N_{k}$, results in minimal variation in the selection of accent subsets. This convergence of test WER implies that as the sample size increases, the specific choice of accent subsets becomes less influential in determining the performance.

\section{Conclusion}

In conclusion, our research addresses the challenge of speech recognition for African accented speech by exploring the linguistic relationships of accent embeddings obtained through AccentFold. Our exploratory analysis of AccentFold provides insights into the spatial relationships between accents and reveals that accent embeddings group together based on geographic and language family similarities, capturing phonological, and morphological regularities based on language families. Furthermore, we reveal, in Section \ref{sec:ethnologue-challenge}, two interesting relationships in some African accents that have been uncharacterized by the Ethnologue. Our experimental setup demonstrates the practicality of AccentFold as an accent subset selection method for adapting ASR models to targeted accents. With a WER improvement of 3.5\%, AccentFold presents a promising approach for improving ASR performance on accented speech, particularly in the context of African accents, where data scarcity and budget constraints pose significant challenges. Our research paves the way for a deeper understanding of accent diversity and linguistic affiliations, thereby opening new avenues for leveraging linguistic knowledge in adapting ASR systems to target accents.
\section*{Limitations}
One limitation of our study is the utilization of a single pre-trained model for fine-tuning in our experiments. While the chosen model demonstrated promising performance, this approach may the generalizability and robustness of our findings. Incorporating multiple pre-trained models with varying architectures and configurations would provide a more comprehensive evaluation of the ASR system's performance.

Furthermore, our study primarily focuses on improving the ASR performance for English with a focus on African accents. Consequently, the findings and outcomes may not be directly transferable to languages outside of the African continent. The characteristics and phonetic variations inherent in non-African accents require tailored approaches to improve ASR systems in different linguistic contexts. Future studies should expand the scope to encompass a broader range of languages and accents to enhance the generalizability of our method beyond African languages.

t-SNE, a stochastic dimensionality reduction algorithm, is highly effective in preserving local structures and representing non-linear relationships in data \cite{ROCA2023100390}. Hence it serves as a versatile and robust tool for visualizing high-dimensional data and has been used extensively in myriad domains: for example in the medical domain it is used in visualizing and understanding single-cell sequencing data \cite{Becht2019dimensionality,Kobak2019art}. However, it should be noted that t-SNE is primarily used for data visualization purposes. Therefore, the insights discussed in Section \ref{sec:ling-analyis} are solely derived from the exploratory analysis conducted using AccentFold and are not based on the inherent capabilities of t-SNE itself. The results obtained from t-SNE analysis should be interpreted with caution, as previous research has demonstrated \cite{ROCA2023100390,Becht2018}.

\section*{Ethics Statement}

We use AfriSpeech-200 dataset \cite{olatunji2023afrispeech200} in this paper to run our experiments. This dataset is released under CC BY-NC-SA 4.0. As we use it only for research purpose or not for any commercial purpose, we do not go against the license. We do not foresee any harmful effects or usages of the methodology proposed or the models. We release all the artefacts created as part of this work under CC BY-NC-SA 4.0.

\bibliography{anthology,custom}
\bibliographystyle{acl_natbib}

\appendix


\begin{figure*}[h]

\includegraphics[width=\textwidth]{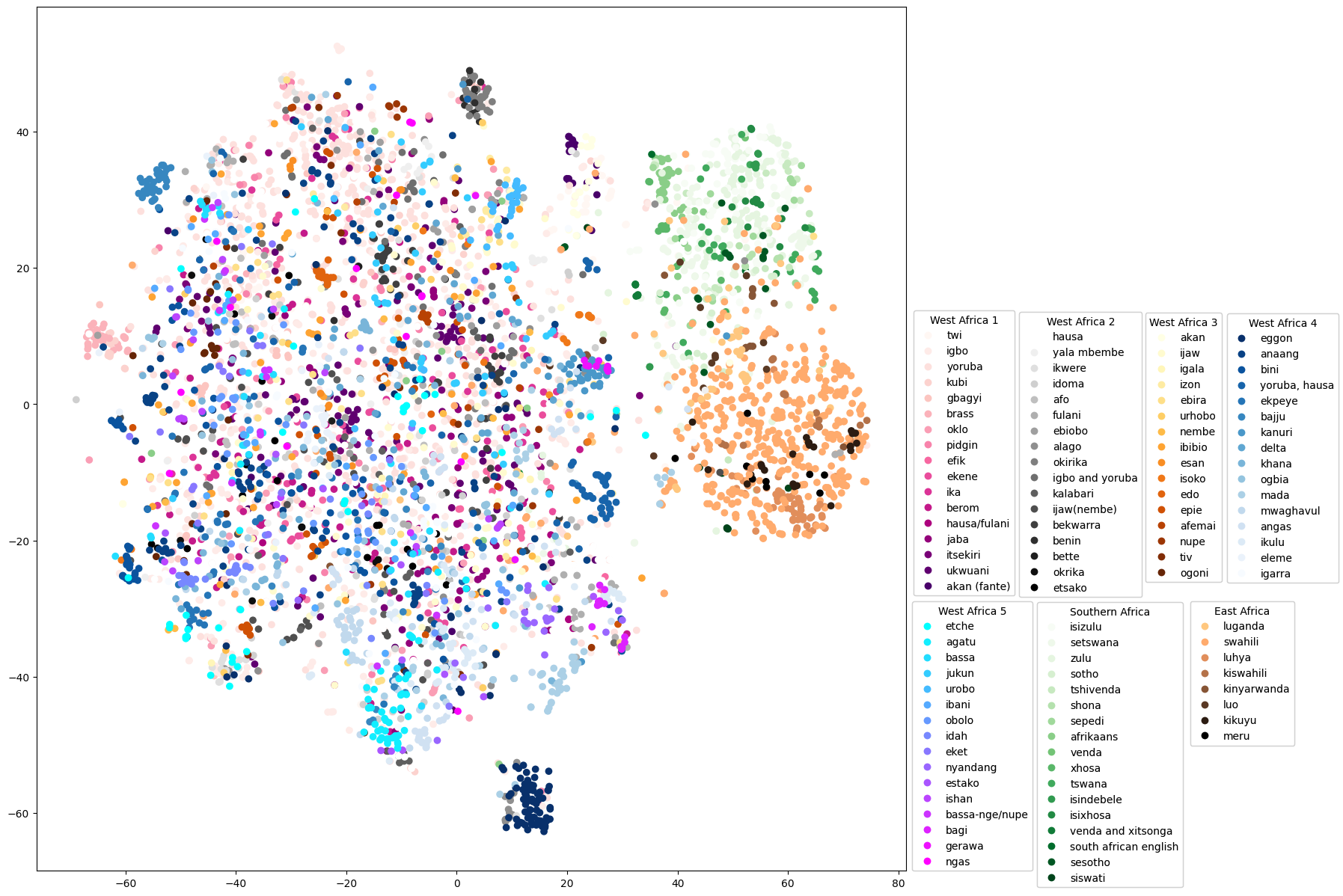} %
\caption{Clustering of Afrispeech test split by Accent}
\label{afrispeech_test_all_data_by_accent}
\end{figure*}


\begin{figure*}[h]
\includegraphics[width=\textwidth]{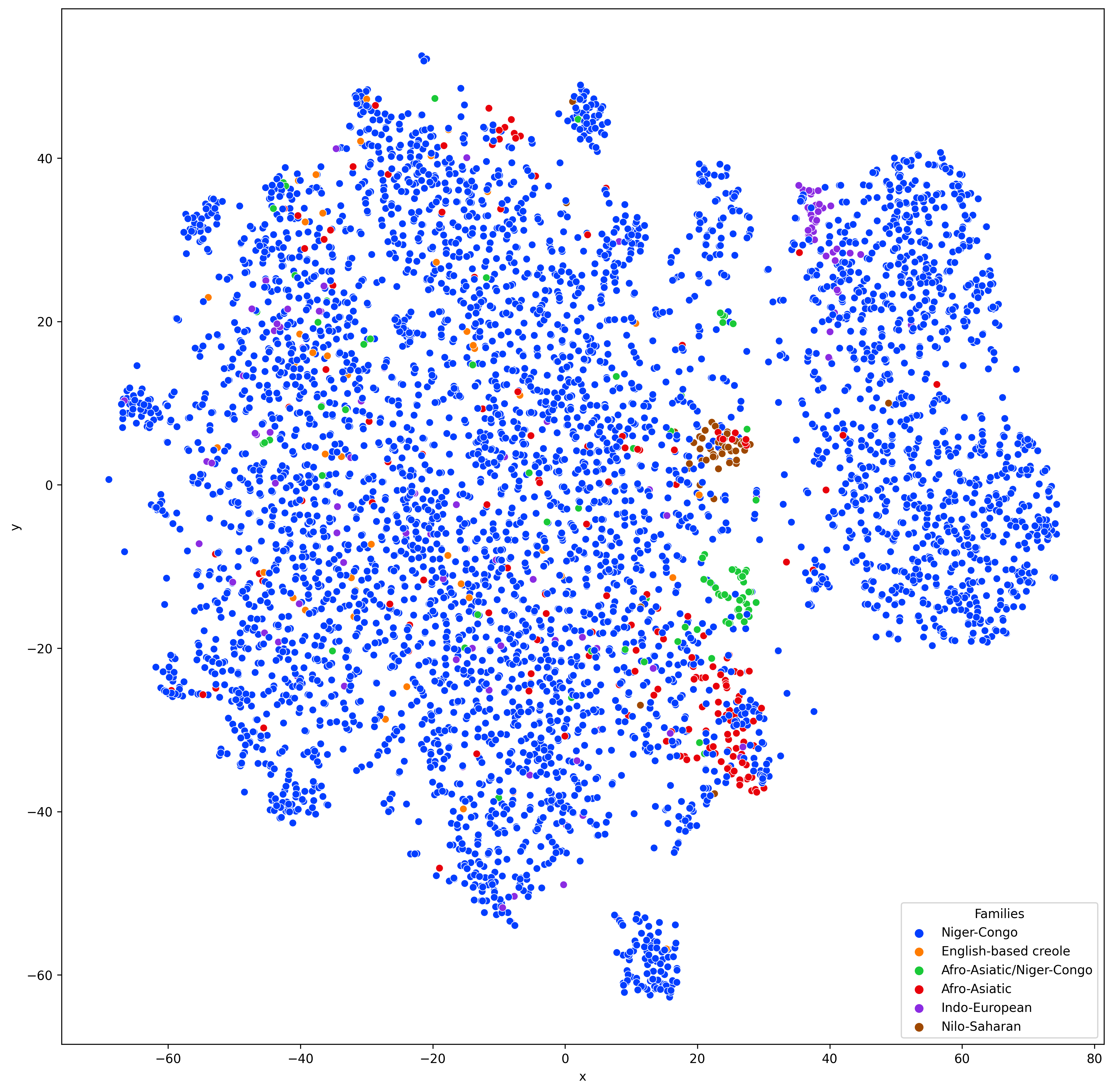} %
\caption{Clustering of Afrispeech test split by language families}
\label{afrispeech_test_all_data_by_families}
\end{figure*}

\begin{figure*}[h]
\centering
\includegraphics[width=19cm]{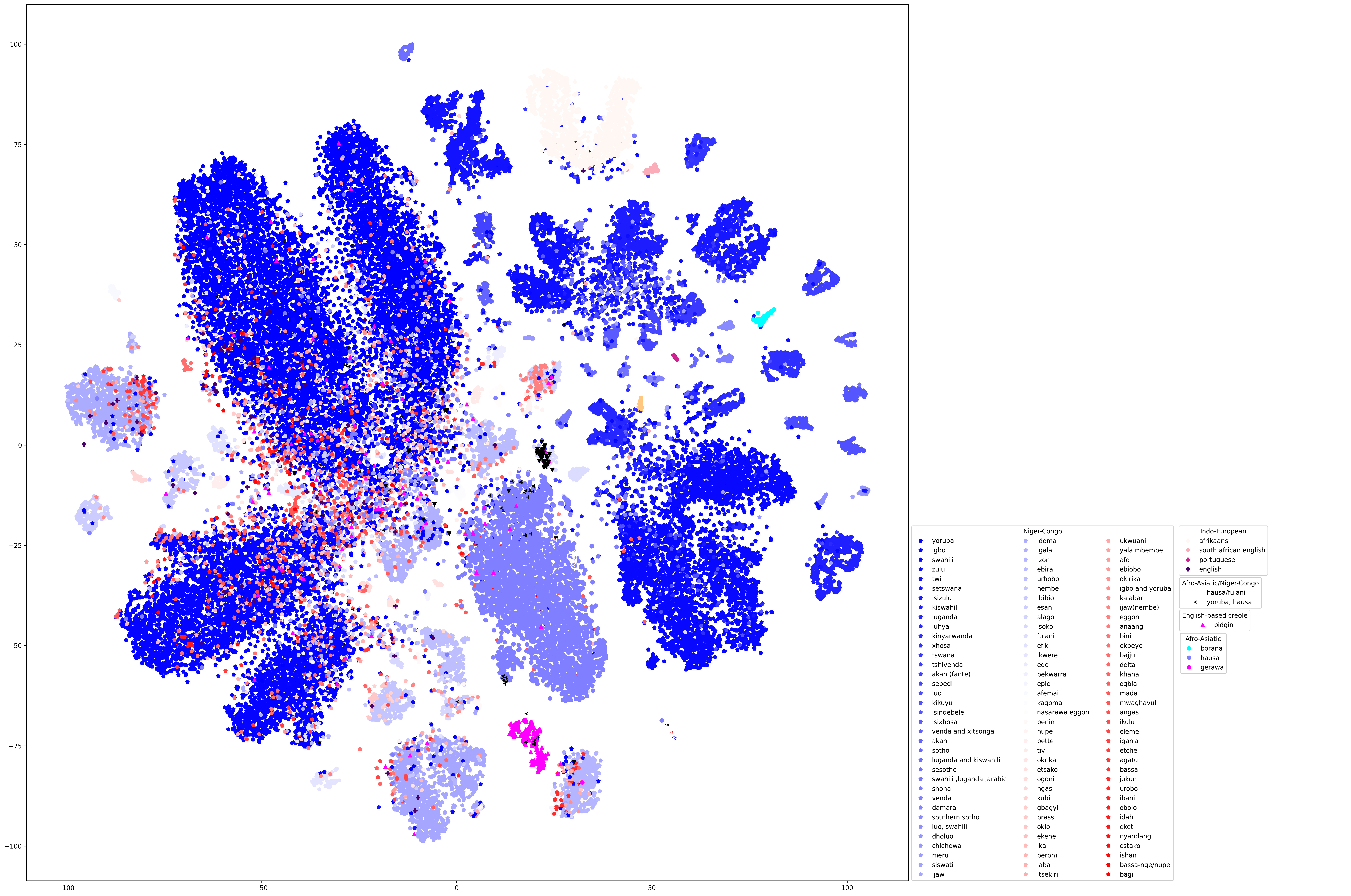} %
\caption{Clustering of the entire Afrispeech data by language families}
\label{afrispeech_train_dev_test_by_family}
\end{figure*}

\begin{figure*}[h]
\includegraphics[width=\textwidth]{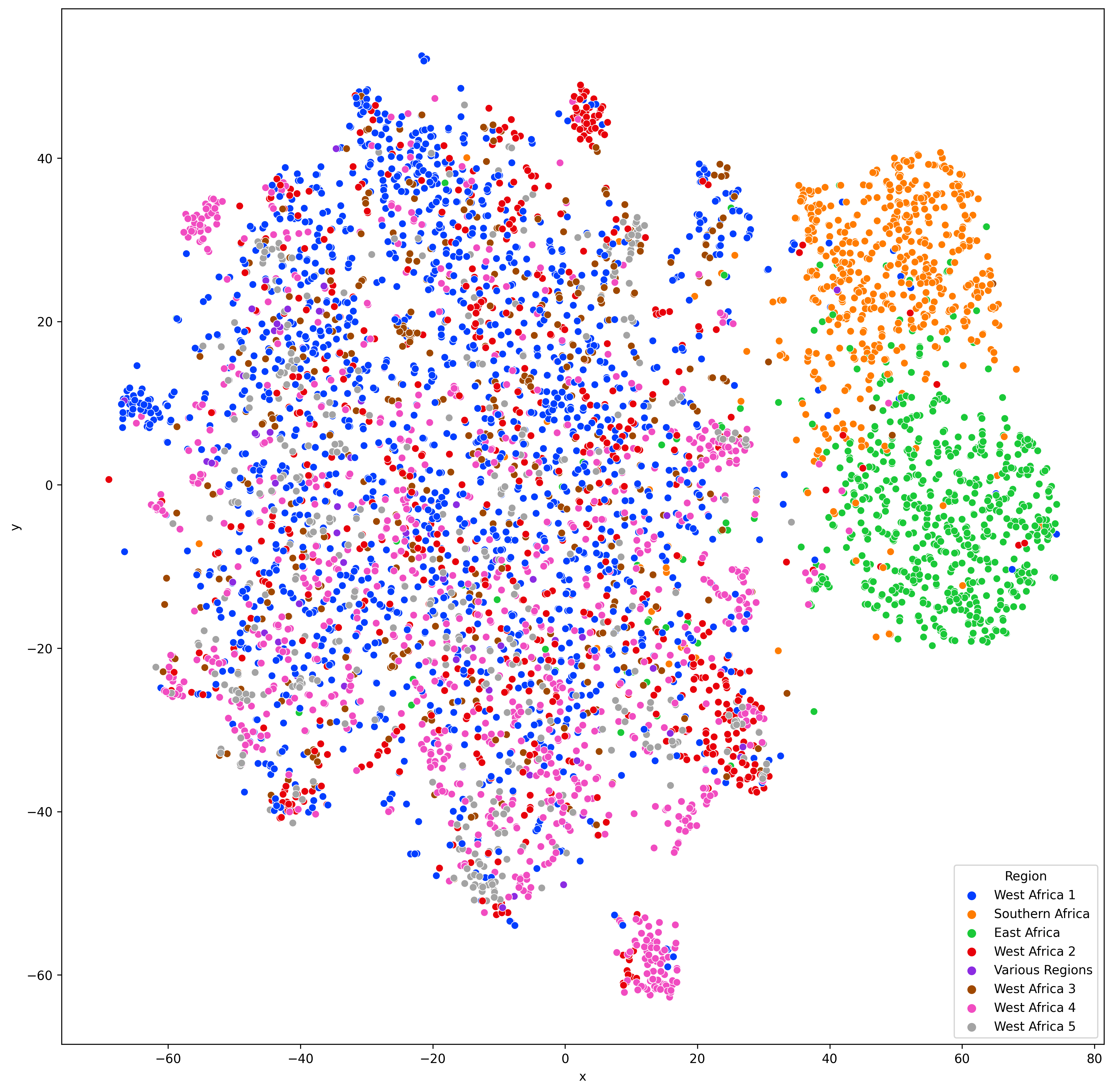}
\caption{t-SNE visualization of AccentFold by region from the Afrispeech test split} 
\label{afrispeech_by_region}
\end{figure*}

\begin{table*}[htb!]
\caption{Accent statistics of Afrispeech dataset}
\footnotesize
\label{syn-stats}
\vskip 0.15in
\begin{center}
\begin{small}
\resizebox{\columnwidth}{!}{
\begin{tabular}{lcccr}
\toprule
 \textbf{Accent}    & \textbf{Clips}	    & \textbf{Country}    & \textbf{Region}    & \textbf{Family}   \\

\midrule
yoruba	    & 15407	    & US,NG	    & West Africa	    & Niger-Congo   \\
igbo	    & 8677	    & US,NG,ZA	    & West Africa	    & Niger-Congo   \\
swahili	    & 6320	    & KE,TZ,ZA,UG	    & East Africa	    & Niger-Congo   \\
hausa	    & 5765	    & NG	    & West Africa	    & Afro-Asiatic   \\
ijaw	    & 2499	    & NG	    & West Africa	    & Niger-Congo   \\
afrikaans	    & 2048	    & ZA	    & Southern Africa	    & Indo-European   \\
idoma	    & 1877	    & NG	    & West Africa	    & Niger-Congo   \\
zulu	    & 1794	    & ZA,TR,LS	    & Southern Africa	    & Niger-Congo   \\
setswana	    & 1588	    & BW,ZA	    & Southern Africa	    & Niger-Congo   \\
twi	    & 1566	    & GH	    & West Africa	    & Niger-Congo   \\
isizulu	    & 1048	    & ZA	    & Southern Africa	    & Niger-Congo   \\
igala	    & 919	    & NG	    & West Africa	    & Niger-Congo   \\
izon	    & 838	    & NG	    & West Africa	    & Niger-Congo   \\
kiswahili	    & 827	    & KE	    & East Africa	    & Niger-Congo   \\
ebira	    & 757	    & NG	    & West Africa	    & Niger-Congo   \\
luganda	    & 722	    & UG,BW,KE	    & East Africa	    & Niger-Congo   \\
urhobo	    & 646	    & NG	    & West Africa	    & Niger-Congo   \\
nembe	    & 578	    & NG	    & West Africa	    & Niger-Congo   \\
ibibio	    & 570	    & NG	    & West Africa	    & Niger-Congo   \\
pidgin	    & 514	    & NG	    & West Africa	    & English-based creole   \\
luhya	    & 508	    & KE	    & East Africa	    & Niger-Congo   \\
kinyarwanda	    & 469	    & RW	    & East Africa	    & Niger-Congo   \\
xhosa	    & 392	    & ZA	    & Southern Africa	    & Niger-Congo   \\
tswana	    & 387	    & ZA,BW	    & Southern Africa	    & Niger-Congo   \\
esan	    & 380	    & NG	    & West Africa	    & Niger-Congo   \\
alago	    & 363	    & NG	    & West Africa	    & Niger-Congo   \\
tshivenda	    & 353	    & ZA	    & Southern Africa	    & Niger-Congo   \\
fulani	    & 312	    & NG	    & West Africa	    & Niger-Congo   \\
isoko	    & 298	    & NG	    & West Africa	    & Niger-Congo   \\
akan (fante)	    & 295	    & GH	    & West Africa	    & Niger-Congo   \\
ikwere	    & 293	    & NG	    & West Africa	    & Niger-Congo   \\
sepedi	    & 275	    & ZA	    & Southern Africa	    & Niger-Congo   \\
efik	    & 269	    & NG	    & West Africa	    & Niger-Congo   \\
edo	    & 237	    & NG	    & West Africa	    & Niger-Congo   \\
luo	    & 234	    & UG,KE	    & East Africa	    & Niger-Congo   \\
kikuyu	    & 229	    & KE	    & East Africa	    & Niger-Congo   \\
bekwarra	    & 218	    & NG	    & West Africa	    & Niger-Congo   \\
isixhosa	    & 210	    & ZA	    & Southern Africa	    & Niger-Congo   \\
hausa/fulani	    & 202	    & NG	    & West Africa	    & Afro-Asiatic/Niger-Congo   \\
epie	    & 202	    & NG	    & West Africa	    & Niger-Congo   \\
isindebele	    & 198	    & ZA	    & Southern Africa	    & Niger-Congo   \\
venda and xitsonga	    & 188	    & ZA	    & Southern Africa	    & Niger-Congo   \\
sotho	    & 182	    & ZA	    & Southern Africa	    & Niger-Congo   \\
akan	    & 157	    & GH	    & West Africa	    & Niger-Congo   \\
nupe	    & 156	    & NG	    & West Africa	    & Niger-Congo   \\
anaang	    & 153	    & NG	    & West Africa	    & Niger-Congo   \\
english	    & 151	    & NG	    & Various Regions	    & Indo-European   \\
afemai	    & 142	    & NG	    & West Africa	    & Niger-Congo   \\
shona	    & 138	    & ZA,ZW	    & Southern Africa	    & Niger-Congo   \\
eggon	    & 137	    & NG	    & West Africa	    & Niger-Congo   \\
luganda and kiswahili	    & 134	    & UG	    & East Africa	    & Niger-Congo   \\
ukwuani	    & 133	    & NG	    & West Africa	    & Niger-Congo   \\
sesotho	    & 132	    & ZA	    & Southern Africa	    & Niger-Congo   \\
benin	    & 124	    & NG	    & West Africa	    & Niger-Congo   \\
kagoma	    & 123	    & NG	    & West Africa	    & Niger-Congo   \\
nasarawa eggon	    & 120	    & NG	    & West Africa	    & Niger-Congo   \\
tiv	    & 120	    & NG	    & West Africa	    & Niger-Congo   \\
south african english	    & 119	    & ZA	    & Southern Africa	    & Indo-European   \\
borana	    & 112	    & KE	    & East Africa	    & Afro-Asiatic   \\
swahili ,luganda ,arabic	    & 109	    & UG	    & East Africa	    & Niger-Congo   \\
ogoni	    & 109	    & NG	    & West Africa	    & Niger-Congo   \\
mada	    & 109	    & NG	    & West Africa	    & Niger-Congo   \\
bette	    & 106	    & NG	    & West Africa	    & Niger-Congo   \\
berom	    & 105	    & NG	    & West Africa	    & Niger-Congo   \\
bini	    & 104	    & NG	    & West Africa	    & Niger-Congo   \\
ngas	    & 102	    & NG	    & West Africa	    & Niger-Congo   \\
etsako	    & 101	    & NG	    & West Africa	    & Niger-Congo   \\
okrika	    & 100	    & NG	    & West Africa	    & Niger-Congo   \\
venda	    & 99	    & ZA	    & Southern Africa	    & Niger-Congo   \\
siswati	    & 96	    & ZA	    & Southern Africa	    & Niger-Congo   \\
damara	    & 92	    & NG	    & Southern Africa	    & Niger-Congo   \\
yoruba, hausa	    & 89	    & NG	    & West Africa	    & Afro-Asiatic/Niger-Congo   \\
southern sotho	    & 89	    & ZA	    & Southern Africa	    & Niger-Congo   \\
kanuri	    & 86	    & NG	    & West Africa	    & Nilo-Saharan   \\
itsekiri	    & 82	    & NG	    & West Africa	    & Niger-Congo   \\
ekpeye	    & 80	    & NG	    & West Africa	    & Niger-Congo   \\
mwaghavul	    & 78	    & NG	    & West Africa	    & Niger-Congo   \\
bajju	    & 72	    & NG	    & West Africa	    & Niger-Congo   \\
luo, swahili	    & 71	    & KE	    & East Africa	    & Niger-Congo   \\
dholuo	    & 70	    & KE	    & East Africa	    & Niger-Congo   \\
ekene	    & 68	    & NG	    & West Africa	    & Niger-Congo   \\
jaba	    & 65	    & NG	    & West Africa	    & Niger-Congo   \\
ika	    & 65	    & NG	    & West Africa	    & Niger-Congo   \\
angas	    & 65	    & NG	    & West Africa	    & Niger-Congo   \\
ateso	    & 63	    & UG	    & East Africa	    & Nilo-Saharan   \\
brass	    & 62	    & NG	    & West Africa	    & Niger-Congo   \\
ikulu	    & 61	    & NG	    & West Africa	    & Niger-Congo   \\
eleme	    & 60	    & NG	    & West Africa	    & Niger-Congo   \\
chichewa	    & 60	    & MW	    & Southern Africa	    & Niger-Congo   \\
oklo	    & 58	    & NG	    & West Africa	    & Niger-Congo   \\
meru	    & 58	    & KE	    & East Africa	    & Niger-Congo   \\
agatu	    & 55	    & NG	    & West Africa	    & Niger-Congo   \\
okirika	    & 54	    & NG	    & West Africa	    & Niger-Congo   \\
igarra	    & 54	    & NG	    & West Africa	    & Niger-Congo   \\
ijaw(nembe)	    & 54	    & NG	    & West Africa	    & Niger-Congo   \\
khana	    & 51	    & NG	    & West Africa	    & Niger-Congo   \\
ogbia	    & 51	    & NG	    & West Africa	    & Niger-Congo   \\
gbagyi	    & 51	    & NG	    & West Africa	    & Niger-Congo   \\
portuguese	    & 50	    & ZA	    & Various Regions	    & Indo-European   \\
delta	    & 49	    & NG	    & West Africa	    & Niger-Congo   \\
bassa	    & 49	    & NG	    & West Africa	    & Niger-Congo   \\
etche	    & 49	    & NG	    & West Africa	    & Niger-Congo   \\
kubi	    & 46	    & NG	    & West Africa	    & Niger-Congo   \\
jukun	    & 44	    & NG	    & West Africa	    & Niger-Congo   \\
igbo and yoruba	    & 43	    & NG	    & West Africa	    & Niger-Congo   \\
urobo	    & 43	    & NG	    & West Africa	    & Niger-Congo   \\
kalabari	    & 42	    & NG	    & West Africa	    & Niger-Congo   \\
ibani	    & 42	    & NG	    & West Africa	    & Niger-Congo   \\
obolo	    & 37	    & NG	    & West Africa	    & Niger-Congo   \\
idah	    & 34	    & NG	    & West Africa	    & Niger-Congo   \\
bassa-nge/nupe	    & 31	    & NG	    & West Africa	    & Niger-Congo   \\
yala mbembe	    & 29	    & NG	    & West Africa	    & Niger-Congo   \\
eket	    & 28	    & NG	    & West Africa	    & Niger-Congo   \\
afo	    & 26	    & NG	    & West Africa	    & Niger-Congo   \\
ebiobo	    & 25	    & NG	    & West Africa	    & Niger-Congo   \\
nyandang	    & 25	    & NG	    & West Africa	    & Niger-Congo   \\
ishan	    & 23	    & NG	    & West Africa	    & Niger-Congo   \\
bagi	    & 20	    & NG	    & West Africa	    & Niger-Congo   \\
estako	    & 20	    & NG	    & West Africa	    & Niger-Congo   \\
gerawa	    & 13	    & NG	    & West Africa	    & Afro-Asiatic   \\

\bottomrule

\end{tabular}
}
\end{small}
\end{center}
\vskip -0.1in
\end{table*}

\end{document}